%% file: main.tex
\pdfoutput=1

\documentclass[11pt]{article}
\usepackage[table]{xcolor} %
\usepackage{adjustbox} %
\usepackage{listings}
\usepackage{multirow}
\usepackage{enumitem}
\usepackage{pifont} %
\usepackage{mathrsfs}
\usepackage{amssymb}
\usepackage{amsmath}
\usepackage{xcolor}
\usepackage{xspace}
\usepackage{booktabs}
\usepackage{subcaption}

\usepackage[final]{acl}

\usepackage{times}
\usepackage{latexsym}

\usepackage[T1]{fontenc}

\usepackage[utf8]{inputenc}

\usepackage{microtype}

\usepackage{inconsolata}

\usepackage{graphicx}

\title{Maximal Matching Matters: Preventing Representation Collapse for Robust Cross-Modal Retrieval}

\author{Hani Alomari \\
  Virginia Tech \\
  hani@vt.edu \\\And
  Anushka Sivakumar \\
  Virginia Tech \\
  anushkas01@vt.edu \\\And
  Andrew Zhang \\
  Virginia Tech \\
  azhang42@vt.edu \\\And
  Chris Thomas \\
  Virginia Tech \\
  chris@cs.vt.edu }

\makeatletter
\DeclareRobustCommand\onedot{\futurelet\@let@token\@onedot}
\def\@onedot{\ifx\@let@token.\else.\null\fi\xspace}

\def\eg{\emph{e.g}\onedot}

\def\etc{\emph{etc}\onedot}

\def\ours{MaxMatch }
\def\oursnospace{MaxMatch}

\makeatother

\newcommand{\cmark}{\ding{51}} %
\newcommand{\xmark}{\ding{55}} %

\usepackage{listings}
\usepackage{xcolor}

\definecolor{keywordcolor}{rgb}{0.5,0,0.5}
\definecolor{stringcolor}{rgb}{0.2,0.6,0.2}
\definecolor{commentcolor}{rgb}{0.4,0.4,0.4}
\definecolor{bgcolor}{rgb}{0.95,0.95,0.95}

\begin{document}
\maketitle
\input{sec/0_abstract}
\input{sec/1_intro}

\input{sec/2_related_works}

\input{sec/3_methodology}

\input{sec/4_experiment}
\input{sec/5_conclusion}

\nocite{andrew2007scalable}
\bibliography{custom}

\appendix

\label{sec:appendix}

\input{sec/supp}
\end{document}

%% file: sec/0_abstract.tex
\begin{abstract}
    Cross-modal image-text retrieval is challenging because of the diverse possible associations between content from different modalities. 
    Traditional methods learn a single-vector embedding to represent semantics of each sample, but struggle to capture nuanced and diverse relationships that can exist across modalities. Set-based approaches, which represent each sample with multiple embeddings, offer a promising alternative, as they can capture richer and more diverse  relationships. 
    In this paper, we show that, despite their promise, these set-based representations continue to face issues including sparse supervision and set collapse, which limits their effectiveness. 
    To address these challenges, we propose \textbf{Maximal Pair Assignment Similarity} to optimize one-to-one matching between embedding sets which preserve semantic diversity within the set. We also introduce two loss functions to further enhance the representations: \textbf{Global Discriminative Loss} to enhance distinction among embeddings, and \textbf{Intra-Set Divergence Loss} to prevent collapse within each set. Our method achieves state-of-the-art performance on MS-COCO and Flickr30k without relying on external data.
\end{abstract}

%% file: sec/1_intro.tex
\section{Introduction}
\label{sec:intro}
Cross-modal retrieval methods aim to align different modalities, such as images and text, by learning shared semantic representations \cite{DeVISE}. The challenge lies in bridging the semantic gap between modalities: each modality contributes unique information, and aligning them requires identifying which features to preserve or discard. Moreover, the relationship between modalities is often one-to-many (\eg, a single image can be described by multiple captions, each emphasizing different aspects). Handling this complexity requires learning robust embeddings that capture the diverse semantic relationships that exist across modalities \cite{song2019polysemousvisualsemanticembeddingcrossmodal, chun2021probabilisticembeddingscrossmodalretrieval}.

\input{figures/fig_1}

Most existing approaches rely on learning a single shared embedding space for images and text \cite{vse, HREM_2023_CVPR, cora2024} performing well on benchmarks like Flickr30k and COCO \cite{flickr30k, cococommonobjects} but struggling with abstract or nuanced cross-modal relationships. To address this, some methods use cross-attention networks that predict similarity between images and text by attending to both modalities simultaneously \cite{cross_attn_nw_wei2020, Diao_Zhang_Ma_Lu_2021, lee2018stackedcrossattentionimagetext, miech2021thinkingfastslowefficient}, yet are computationally intensive, requiring joint processing of image-text pairs for each query, which limits their scalability. A common approach is to use separate visual and textual encoders, which allow for efficient search on precomputed embeddings \cite{gu2018lookimaginematchimproving, visualtextalignment_messina2021}. This method scales well to large datasets but often fails to capture the diversity of cross-modal relationships because each input is reduced to a single embedding vector limiting the representation's ability to handle one-to-many relationships between queries and potential matches.

A few recent works have proposed architectures that aim to capture these diverse relationships using separate encoders \cite{song2019polysemousvisualsemanticembeddingcrossmodal, chun2023improved, kim2023setdiverseembeddings}. The goal is to reduce ambiguity and improve alignment between modalities by representing multiple facets of each input with a set of embedding vectors. Figure \ref{fig:figure_1} shows how these embedding sets cluster similar features from both modalities in a shared space. 

While set-based embedding models offer a promising solution for capturing diverse cross-modal relationships, the loss and similarity functions to train these methods lead to issues like sparse supervision (where some embeddings are unused) and set collapse (where embeddings collapse into a single point). For example, PVSE \cite{song2019polysemousvisualsemanticembeddingcrossmodal} uses a distance function that relies on maximum similarity within the set, leaving many embeddings undertrained. Our analysis shows that Smooth-Chamfer similarity in SetDiv \cite{kim2023setdiverseembeddings}, proposed as a solution to this issue, leads to low variance across the embeddings within the set, reducing the ability to capture diverse relationships in the data.

 In this paper, we address these limitations by introducing \oursnospace, a new Maximal Pair Assignment Similarity mechanism and diversity-promoting losses. Unlike previous work, our approach optimally matches pairs within the embedding sets using a \emph{permutation-based} similarity, ensuring that every embedding in the set contributes to the final objective. In addition, we design a global discriminative loss that encourages each embedding to diverge from a global reference vector, and an intra-set divergence constraint that pushes embeddings apart within the set. These strategies effectively prevent set collapse and improve alignment across modalities. Specifically, our contributions include:

\begin{itemize}[noitemsep,leftmargin=*]
    \item We introduce a new similarity measure for measuring the distance between sets of embeddings. \ours uses permutation-based assignments and the Hungarian algorithm to match embeddings between sets and compute their distance. We show that \ours prevents degenerate embedding sets unlike previous methods.

    \item We propose a novel loss which enforces a margin between residual embeddings and a global reference embedding to encourage semantic diversity. We also introduce intra-set divergence constraints to further discourage set collapse. We show that these functions promote semantic diversity and improve the model's discriminative capability.

    \item We conduct a comprehensive experimental analysis of our proposed technique against a number of state-of-the-art methods, including embedding space visualizations and analysis. Our results show that \ours learns richer, more diverse embedding sets while outperforming similar methods.
\end{itemize}

%% file: figures/fig_1.tex
\begin{figure}[t]
  \centering
  \includegraphics[width=1\linewidth, height=2.25in]{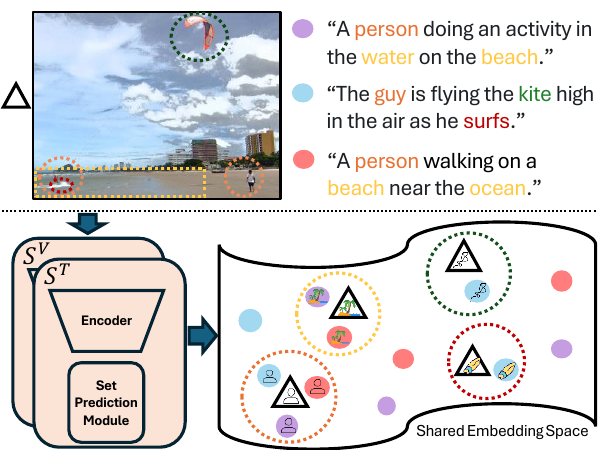}

   \caption{ We illustrate how the same image can be paired with multiple captions. 
   Color-coded regions link visual elements to corresponding textual descriptions. 
   Our method produces multiple embeddings for each sample (image or caption) in a shared space where similar visual and textual elements cluster (indicated by colored groups and dashed boundaries). Icons on embeddings illustrate the model's capacity to map multiple image regions to relevant text descriptions. }
   \label{fig:figure_1}

\end{figure}

%% file: sec/2_related_works.tex
\input{figures/fig_2}

\vspace{-1em}

\section{Related Work}

\noindent\textbf{Cross-modal Retrieval:} Methods fall into two main categories: independent-embedding and interactive-embedding approaches. Independent-embedding approaches use separate encoders for each modality to learn a shared embedding space \cite{vse, deep_corr_yan2015, gu2018lookimaginematchimproving}, enabling efficient retrieval through precomputed embeddings. While researchers have explored various improvements to similarity metrics \cite{wei2020universalweightingmetriclearning, kim2023setdiverseembeddings}, loss functions \cite{thomas2020preserving, chun2021probabilisticembeddingscrossmodalretrieval}, and architectures \cite{gu2018lookimaginematchimproving}, these methods still struggle with complex cross-modal relationships \cite{song2019polysemousvisualsemanticembeddingcrossmodal}. Recent work like CORA \cite{cora2024} and 3SHNet \cite{ge20243shnet} achieve improvements by incorporating external knowledge (scene graphs, segmentation), but this makes direct comparisons with methods that do not use external data more difficult. 

Interactive-embedding approaches instead use cross-attention networks for direct similarity estimation \cite{cross_attn_nw_wei2020, Diao_Zhang_Ma_Lu_2021}, particularly when they are built on large pretrained architectures like CLIP \cite{radford2021learning}. While effective, these methods are computationally intensive as they require joint processing of each query-candidate pair at inference time. In contrast, \ours focuses on improving independent embeddings to maintain efficiency while better capturing complex relationships.

\noindent\textbf{Multi-embedding Representations:}
Recently, a few papers have emphasized the limitations of representing each sample as a single embedding vector for cross-modal retrieval \cite{cora2024, vse, chen2021learningbestpoolingstrategy, HREM_2023_CVPR, thomas2022emphasizing}. A single vector often fails to capture the rich and diverse semantics present in the data. One strategy to address this challenge is set-based representation learning, where multiple embeddings are generated for each input, like in SetDiv, PVSE, and PCME \cite{chun2021probabilisticembeddingscrossmodalretrieval}. These approaches aim to reduce ambiguity by capturing the varied semantics of each modality using a set-based embedding representation. SetDiv introduces smooth-Chamfer similarity to overcome sparse supervision issues, but it leads to a new problem of set collapse, where embeddings lose diversity. To address sparse supervision and set collapse, we introduce a Maximal Pair Assignment Similarity function that explicitly preserves diversity. Even though PVSE and SetDiv use diversity loss to encourage distinct representations by penalizing pairwise similarity but lack global context, focusing only on residual embeddings before fusion. To address this, we propose novel pushing losses to enhance intra-set diversity and better capture modality relationships, achieving improved cross-modal retrieval performance compared to prior methods.

%% file: figures/fig_2.tex
\begin{figure*}[t]
  \centering
  \includegraphics[width=1\linewidth, height=2.5in]{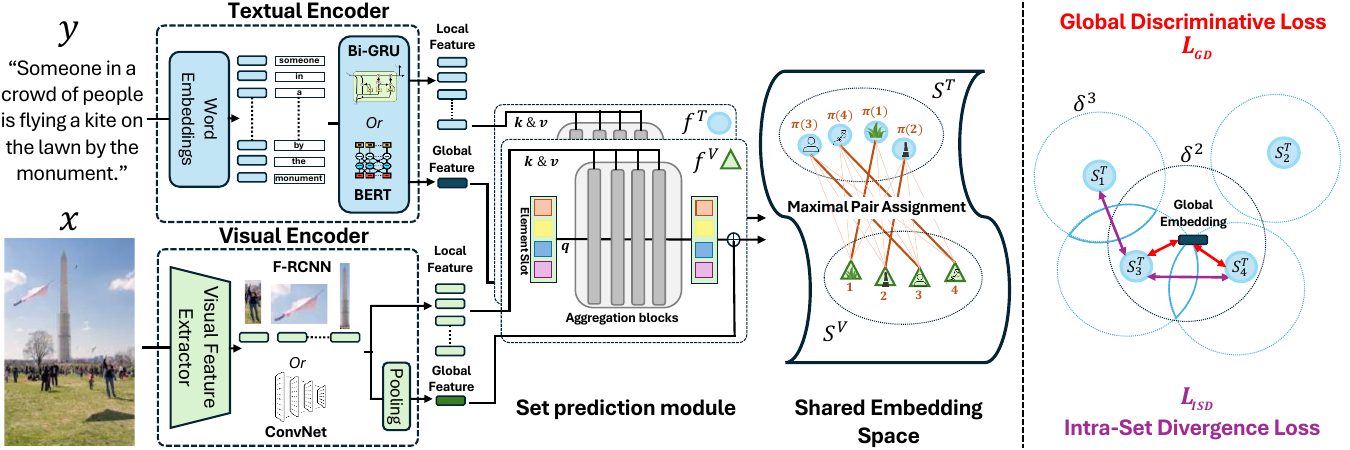}

   \caption{Left: An overview of the model architecture based on \citet{kim2023setdiverseembeddings} consisting of a visual encoder, textual encoder, and set prediction modules for $f^V$ and $f^T$. Local and global features are extracted from each modality and input into a set prediction module to generate embedding sets $S^V$ and $S^T$. We contribute three key components which enable the model to learn diverse embedding sets: a Maximal Pair Assignment Similarity function, Global Discriminative Loss, and Intra-Set Divergence Loss. Right: Our Global Discriminative Loss pushes embeddings within each set away from the global embedding, preventing set collapse, while the Intra-Set Divergence Loss ensures that individual embeddings within each set are distinct, promoting intra-set diversity.}   
   \label{fig:main_figure}
   \vspace{-1em}
\end{figure*}

%% file: sec/3_methodology.tex
\vspace{-0.75em}
\section{Method}
\vspace{-0.5em}
\subsection{Feature Extraction}

\ours architecture consists of two encoders, shown in Figure \ref{fig:main_figure}, each connected to a set prediction model. For vision, a visual encoder $f^{\mathcal{V}}$ takes an image as input $x$ and produces the image embedding vector $v = f^{\mathcal{V}}(x) \in \mathbb{R}$, and a text encoder $f^\mathcal{T}$ that takes in the text caption $y$ and produces its embedding $t = f^\mathcal{T}(y) \in \mathbb{R}$. Each encoder has two branches that compute local features and global features from the input sample. The extracted local features are given as input to the set prediction modules, each of which fuses the local and global features to encode them into an embedding set. We follow the settings of previous works \cite{song2019polysemousvisualsemanticembeddingcrossmodal, kim2023setdiverseembeddings} for the visual and textual feature extractors.

\noindent \textbf{Visual Feature Extractor:} 
We use one of two types of visual feature extractors following previous works like PVSE and SetDiv: (1) Flattened convolutional feature maps as local features $\psi^\mathcal{V}(x) \in \mathbb{R}^{N \times D}$ and their average pooled features as the global feature $\phi^\mathcal{V}(x) \in \mathbb{R}^D$. (2) ROI features from a pretrained object detector as local features $\psi^\mathcal{V}(x) \in \mathbb{R}^{N \times D}$ and their max-pooled representation as the global feature $\phi^\mathcal{V}(x) \in \mathbb{R}^D$.

\noindent \textbf{Textual Feature Extractor:} We use one of two types of text feature extractors: (1) Bi-GRU: An input caption $y$ of $L$ words is encoded using GloVe \cite{glove}, producing 300-dimensional word vectors as local features $\psi^{\mathcal{T}}(y) \in \mathbb{R}^{L \times 300}$. A Bi-GRU with $H$ hidden units processes these features, with the final hidden state representing the global features $\phi^\mathcal{T}(y) \in \mathbb{R}^D$. (2) BERT \cite{devlin2019bertpretrainingdeepbidirectional}: The hidden state outputs serve as local features $\psi^\mathcal{T}(y) \in \mathbb{R}^{L \times D}$, and their max-pooled representation forms the global feature $\phi^\mathcal{T}(y) \in \mathbb{R}^D$.

\input{figures/fig_3}

\subsection{Embedding Set Prediction Module} 
We next describe the set prediction module adapted from \citet{kim2023setdiverseembeddings} that we utilize to generate set embeddings. As shown in Figure \ref{fig:main_figure}, slots in the module iteratively compete to attend to input features. Each of the $L$ aggregation blocks in the module contains a cross-attention layer followed by a feed-forward network. $K$ learnable queries (``slots'') cross-attend to keys and values from $\psi^\mathcal{V}(x)$, denoted $E^l = AggBlock(E^{l-1}) = MLP(\bar{E}^l) + \bar{E}^l \in \mathbb{R}^{K \times D}$, where $\bar{E}^l$ represents the slot states
after cross-attention. The final embedding set $S$ is computed as $S=LN(E^L)+[LN(\phi(y)) ...^{\times K}]$, where $LN$ is a normalization layer, and $[LN(\phi(y)) ...^{\times K}] \in \mathbb{R}^{K \times D}$ denotes $K$ repetitions of global features $\phi(y)$. Conceptually, each slot represents an offset from the global feature $\phi^\mathcal{V}(x)$. However, embeddings in standard SetDiv have a tendency to cluster closely around the global feature (i.e.~set collapse). To encourage diversity, \ours introduces a Maximal Pair Assignment Similarity, Global Discriminative Loss, and Intra-Set Divergence Loss.

\subsection{Maximal Pair Assignment Similarity}
\ours aims to overcome issues related to set-based embeddings and set-based similarity measures, particularly set collapse and sparse supervision. To better understand these limitations, we first examine existing set-based similarity functions such as Multiple Instance Learning (MIL) similarity, introduced in PVSE, and smooth-Chamfer similarity, introduced by SetDiv. 

Given two sets of embeddings $S1 =  \{ x_1,x_2,...,x_K \}$ and $S2 =  \{ y_1,y_2,...,y_K \}$, and a similarity function  $c(x,y)$ between vectors $x \in S_1$ and $y \in S_2$, the MIL loss objective is defined as \(S_{MIL}(S_1,S_2) = \max_{x \in S_1 , y \in S_2} c(x,y)\); while MIL provides a simple mechanism for similarity calculation between embedding sets, it suffers from sparse supervision because the $\max$ function passes gradients to a single element in each set. By focusing only on maximum similarity during training, it underutilizes the remaining elements in the sets and leads to embedding sets where only one element contributes during retrieval. \citet{kim2023setdiverseembeddings} proposed smooth-Chamfer similarity to address the sparse gradient issue. This averages the similarity scores between all pairs from $S_1$ and $S_2$, which allows all embeddings to receive gradients:
\vspace{-1 em}

{
\small
\begin{align}
\label{eqn:sdsdds}
\begin{split}
S_{SC} (S_1,S_2) = \frac{1}{2\alpha|S_1|} \sum_{x \in S_1} \log \big( \sum_{y \in S_2} e^{\alpha c(x,y)} \big) 
\\
+ \frac{1}{2\alpha|S_2|} \sum_{x \in S_2} \log \big( \sum_{y \in S_1} e^{\alpha c(x,y)} \big) 
\end{split}
\end{align} 
}

However, despite providing gradient signals for the entire set, we show in Figures \ref{fig:tsne} and \ref{fig:heatmap} that Smooth-Chamfer similarity leads to set collapse (i.e.~all embeddings become the same), while MIL leads to unused embeddings. Even with a high $\alpha$ value, intended to emphasize stronger similarities, the log function smooths out contributions and leads to uniformly high similarity scores across all elements (i.e.~collapse). See our Appendix for additional analysis. In sum, both existing techniques for training set-based representations lead to degenerate conditions. 
\input{figures/fig_4} 
We propose Maximal Pair Assignment Similarity, which learns optimal matching between elements of embedding sets. By no longer pulling all embeddings towards the mean, sets are no longer encouraged to collapse, yet all embeddings remain active in the matching process.

\ours leverages permutation-based assignments, which we efficiently implement using block processing and masking techniques. It finds the optimal pairing of embeddings from $S_1$ and $S_2$, where the overall matching score of the sets is maximized. We describe our approach in detail below.

\noindent \textbf{Cosine Similarity Calculation:} We begin by computing the cosine similarity between the L2-normalized set of image embeddings $\mathbf{V}_i = \{ v_{i1},v_{i2}, \cdots,v_{iK} \} \in \mathbb{R}^{K \times D}$ and the set of text embeddings $\mathbf{T}_j = \{ t_{j1}, t_{j2}, \cdots, t_{jK} \} \in \mathbb{R}^{K \times D}$, where $K$ represents the number of embeddings per sample. The similarity score between each pair of sets is \( Sim(V_i,T_j) = S(V_i, T_j) \), where $S(V_i, T_j) \in \mathbb{R}^{K \times K}$ is the similarity matrix for the sets: \(S_{mn}(V_i, T_j) = \frac{v_{im} \cdot t_{jn}}{||v_{im}|| ||t_{jn}||} \), where $S_{mn}(V_i, T_j)$ denotes the element in the $m$-th row and $n$-th column of $S(V_i, T_j)$.

\noindent \textbf{Hungarian Algorithm:} To obtain the optimal matching, we apply the Hungarian algorithm to blocks of the similarity matrix $S(V_i, T_j)$ between embedding sets $V_i$ and $T_j$ which yields the permutation $\pi$ that maximizes the total similarity by selecting the most semantically meaningful pairs:

{\small
\begin{align}
\pi^{*} = \underset{\pi}{\arg\max} (\operatorname{tr}(S(V_i, \pi(T_j)))) \label{eqn:permutation}
\end{align}}
\noindent where $\pi$ is a permutation of indices representing the optimal matching between the image and text embeddings and $tr$ denotes trace. After that, we apply the optimal assignment on $S_{ij}$ that maximizes the total similarity, where the outputs are a binary mask $M_{ij}$ indicating the selected matches:

{\small
\begin{align*}
M_{ij} = 
\begin{cases} 
1, & \text{if } (V_i, \pi^* (T_j)) \text{ is an optimal match} \\
0, & \text{otherwise}
\end{cases}
\end{align*}}

\noindent The mask $M_{ij}$ is applied to the similarity matrix to extract only the optimal matches:

\begin{align}
    Max Sim(S_{i}^{\mathcal{V}}, S_{j}^{\mathcal{T}})  = \sum_{k=1}^{K} M_{ij} \odot S_{ij}  \label{max-similarity}
\end{align}
where $M_{ij}$ is a binary mask matrix that indicates whether a particular pair $(V_{i}, \pi^*(T_j))$ is part of the optimal one-to-one matching between the sets $S_{i}^{\mathcal{V}}$ and $S_{j}^{\mathcal{T}}$. Finally, 
we scale and sum them to calculate the final similarity score $S_H$ for each image-text pair:

{\small
\scriptsize
\begin{align}
S_{H} (S_{i}^{\mathcal{V}}, S_{j}^{\mathcal{T}}) = \frac{1}{K} \sum_{i=0}^{K} \sum_{j=0}^{K} \left( \left( \exp \left( MaxSim(S_{i}^{\mathcal{V}}, S_{j}^{\mathcal{T}}) \right) \right) - 1\right) \label{eqn:hungarian-similarity}
\end{align}
}
\par
\noindent where the exponential function $\exp$ is used to amplify the influence of pairs with higher similarity. This approach aligns with the intuition that the closest pairs, those that best capture the semantic relationship between the image and text, should have more weight in the final similarity score. For example, if a particular image-text pair is closely related, their similarity score after exponential scaling will dominate, ensuring that these critical pairs contribute more significantly to the overall score.

\subsection{Training and Inference}
We train \ours to minimize standard objectives presented in previous works \cite{kim2023setdiverseembeddings, cora2024, song2019polysemousvisualsemanticembeddingcrossmodal}: triplet loss (TRI), diversity regularizer (Div), Maximum Mean Discrepancy (MMD) regularizer, and Contrastive loss (CON). In addition, we introduce a Global Discriminative Loss (GD) and Intra-Set Divergence Loss (ISD) to prevent set collapse and encourage the model to learn diverse embeddings. 

{\small
{\begin{align}
\label{eqn:combined-loss}
\begin{split}
  \mathcal{L} = \mathcal{L}_{TRI} + \lambda_{GD} \mathcal{L}_{GD} + \lambda_{ISD} \mathcal{L}_{ISD}
\\
 + \lambda_{MMD} \mathcal{L}_{MMD} + \lambda_{Div} \mathcal{L}_{Div} + \lambda_{CON} \mathcal{L}_{CON}
\end{split}
\end{align} 
}}

\noindent \textbf{Triplet loss with hardest negatives}: Following prior work \cite{kim2023setdiverseembeddings, song2019polysemousvisualsemanticembeddingcrossmodal}, we incorporate the a hinge-based Triplet Loss with Hardest Negatives mining. Given a batch of embeddings \( B = \{ (S_i^\mathcal{V}, S_i^\mathcal{T}) \}_{i=1}^N \), the triplet loss is formulated as follows:

{\small
\begin{align}
\label{eqn:triplet-loss}
\begin{split}
 \mathcal{L}_{TRI} = \sum_{i=1}^{N} \max_{j} \left[ \delta_1 + S_{H}(S_{i}^{\mathcal{V}}, S_{j}^{\mathcal{T}}) \right] - S_{H}(S_{i}^{\mathcal{V}}, S_{i}^{\mathcal{T}}) \bigg]_+
\\
 + \sum_{i=1}^{N} \max_{j} \left[ \delta_1 + S_{H}(S_{i}^{\mathcal{T}}, S_{j}^{\mathcal{V}}) \right] - S_{H}(S_{i}^{\mathcal{T}}, S_{i}^{\mathcal{V}}) \bigg]_+
\end{split}
\end{align}} \\[-1em]
\noindent where \(\delta_1 > 0\) denotes the margin. For each positive pair \( (S_i^\mathcal{V}, S_i^\mathcal{T}) \), the model identifies the hardest negative sample \( S_j^\mathcal{T} \) closest to \( S_i^\mathcal{V} \), and similarly the hardest negative \( S_j^\mathcal{V} \) closest to \( S_i^\mathcal{T} \). %

\noindent \textbf{Regularization and Stabilization Losses:} To ensure robust and stable training, we employ a set of losses following prior work \cite{kim2023setdiverseembeddings}. The Maximum Mean Discrepancy (MMD) \cite{gretton2006kernel} minimizes the MMD between embedding sets of images and text, helping to prevent early divergence between modalities. The diversity regularizer \cite{song2019polysemousvisualsemanticembeddingcrossmodal} penalizes similar element slots. Consistent with previous findings \cite{chen2021learningbestpoolingstrategy}, we observed that using the hardest triplet loss can lead to instability during early training. Therefore, following \cite{cora2024}, we also leverage contrastive loss to align all matching image and text representations.

\noindent \textbf{Global Discriminative Loss:} pushes each embedding away from the corresponding global reference embedding (within a margin and with a scaling factor) within each set to ensure the embedding set captures meaningful semantics beyond what is already captured in the global representation.
The key idea is to ensure that embeddings within each set are gradually pushed away from their respective global, encouraging distinct representations within the set. The loss is defined as

\vspace{-1em}
{\tiny
\begin{align}
\label{eqn:global-discriminative-loss}
\begin{split}
\mathcal{L}_{GD} = \frac{1}{2N} \sum_{i=1}^{N} & \left( \exp \left( s \cdot \left( \cos(v_i, \phi^\mathcal{V}(y_i)) - \delta_{2} \right) \right) \right. \\
& \left. + \exp \left( s \cdot \left( \cos(t_i, \phi^\mathcal{T}(y_i)) - \delta_{2} \right) \right) \right)
\end{split}
\end{align}}

\noindent where \( N \) is the batch size, \( \delta_2 \) is the separation margin, and \( s \) is the scaling factor that controls gradient smoothness. Here, \( \cos(v_i, \phi^\mathcal{V}(y)) \) and \( \cos(t_i, \phi^T(y)) \) represent the  similarities between each visual embedding \( v_i \) or textual embedding \( t_i \) and their respective global references, \( \phi^\mathcal{V}(y) \) and \( \phi^T(y) \). The exponential function applied to the scaled similarity encourages embeddings to spread apart in the feature space if they exceed the margin \( \delta_2 \), which promotes a well-distributed feature space across both modalities.

\noindent \textbf{Intra-Set Divergence Loss:} is designed to encourage diversity among embeddings \textit{within} each set by penalizing high similarity between embeddings. This loss reduces redundancy within the set to encourage sets to learn unique representations which capture different aspects of the data. The loss is formulated as:

{
\tiny
\begin{align}
\label{eqn:intra-set-divergence-loss}
\begin{split}
\mathcal{L}_{ISD} = \frac{2}{M(M-1)} \sum_{j=1}^{M-1} \sum_{k=j+1}^{M} & \left( \exp \left( s \cdot \left( \cos(v_{i,j}, v_{i,k}) - \delta_{3} \right) \right) \right. \\
& \left. + \exp \left( s \cdot \left( \cos(t_{i,j}, t_{i,k}) - \delta_{3} \right) \right) \right)
\end{split}
\end{align}
}
\noindent where \( M \) is the number of embeddings per set, \( \delta_3 \) is the similarity margin, and \( s \) is a scaling factor that smooths the gradients. The cosine similarity terms \( \cos(v_{i,j}, v_{i,k}) \) and \( \cos(t_{i,j}, t_{i,k}) \) measure the alignment between each pair of non-identical embeddings within the visual and textual sets, respectively. By applying the exponential function with scaling, the loss pushes embeddings further apart in the feature space if they exceed the margin \( \delta_3 \), encouraging a diverse distribution. The normalization factor \( \frac{2}{M(M-1)} \) averages the loss across all non-matching pairs.

\noindent \textbf{Inference:} During inference, we select the $top\text{-}k$ most similar embeddings from the predicted sets without applying the one-to-one matching process used during training. This approach reduces the computational complexity and significantly speeds up the inference phase, making the model efficient while still leveraging the diverse embeddings.

%% file: figures/fig_3.tex
\begin{figure}[t]
  \centering
  \includegraphics[width=1\linewidth]{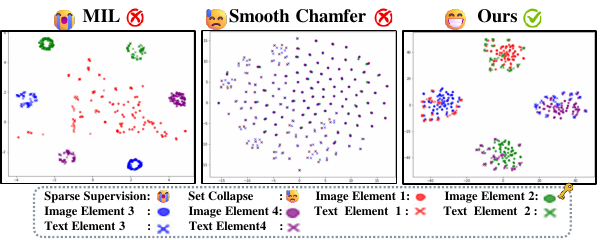}
\vspace{-2em}
   \caption{$t$-SNE visualization of learned embedding spaces, where colors indicate different elements within each set, and markers differentiate modalities (dots for image embeddings, crosses for text embeddings). MIL produces scattered embeddings, Smooth Chamfer collapses them, while \ours preserves semantic distinctiveness and strong alignment.}
   \label{fig:tsne}

\end{figure}

%% file: figures/fig_4.tex
\begin{figure}[t]
  \centering
  \includegraphics[width=1\linewidth]{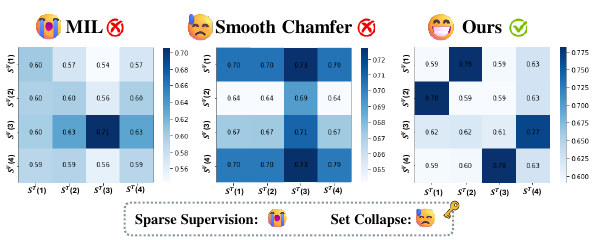}
\vspace{-2em}
   \caption{ Heatmaps showing average pairwise similarities between image-text embeddings across all test samples for models trained with different similarity functions: MIL shows sparse supervision, leading to isolated high similarities (only one embedding in the set is used), Smooth Chamfer causes set collapse, resulting in uniformly high similarities, and \ours maintains distinct alignment, and fully utilizing embeddings set.}
   \label{fig:heatmap}

\end{figure}

%% file: sec/4_experiment.tex
\input{tables/tab2}

\input{tables/tab1}

\section{Experiments}

\subsection{Datasets and Evaluation Metric}
We evaluate \ours on Flickr30k and COCO datasets.
We follow the standard train, validation, test splits and evaluation practices from prior work. 
For COCO, we average performance over five folds of 1K test images and on the full set of 5K test images, while for Flickr30k, we use the test set. In both datasets, each image has five matching captions. Performance is measured using Recall@$K$ ($K \in \{1, 5, 10\}$) and the RSUM metric, which sums Recall@K scores for image-to-text and text-to-image retrieval \cite{vse}. Implementation details are provided in the Appendix.

\subsection{Comparisons with Other Methods}
We report results for the Flickr30k and COCO datasets in tables \ref{table:flickr30k} and \ref{table:mscoco}, respectively. We evaluate our method across two visual feature extractors: ImageNet pretrained ResNet152 \cite{resnet}, and Faster R-CNN with pre-extracted region-of-interest (ROI) features \cite{faster_rcnn}. We use either Bi-GRU or BERT for the text feature extraction. Following PVSE, SetDiv, CORA, \etc, we report ensemble results by averaging similarities from two checkpoints with different seeds. Additionally, the results for the Faster R-CNN + BERT configuration were obtained using the official codebase, as this setting was not reported in the original paper \cite{kim2023setdiverseembeddings}.

We observe that \ours outperforms state-of-the-art methods by an impressive margin except when using Bi-GRU as a semantic concept encoder with the Faster R-CNN visual features. However, \ours achieves the second-best results in this setting. Note that the top-performing method, CORA enhances text representations through a graph generated by an external large language model-based parser \cite{li2023factual}, introducing an additional model and text processing step. Using ResNet152 extracted visual features, \ours substantially outperforms all other methods with a jump in RSUM by $+27.3$ on Flickr30k dataset and an increase in RSUM by $+24.55$ on COCO 5K. With BERT as the semantic concept encoder and Faster-RCNN as the visual feature extractor, \ours outperforms all state-of-the-art methods by a significant margin. In particular, \ours achieves $+2.9$ RSUM improvement over HREM on Flickr30k, and $+1.6$ RSUM on COCO 5K. It is also worth mentioning that \ours outperforms CHAN by $ + 2.3 $ RSUM on Flickr30k,  and $+2.6$ RSUM on COCO 5K which requires cross-attention during inference time, making CHAN more computationally expensive.

\input{tables/tab5}

\subsection{Ablation Study}
Table \ref{table_abl} ablates the different similarity functions (MIL, smooth-Chamfer, and Maximal Pair Assignment) in combination with various loss functions (Div, MMD, GD, and ISD) to analyze their impact on overall performance, as measured by RSUM on the 5k MS-COCO testing dataset. The Maximal Pair Assignment function achieves the highest RSUM score of 446.53 when combined with all four loss settings (Div, MMD, GD, and ISD), resulting in the best performance among all tested configurations. This highlights Maximal Pair Assignment as the most effective similarity function when combined with the appropriate losses. In contrast, other combinations, such as MIL or smooth-Chamfer with selected loss functions, achieve lower RSUM scores, suggesting reduced effectiveness at capturing the desired relationships. Notably, Maximal Pair Assignment with GD and ISD losses exhibits strong performance, achieving RSUM scores above 441, showing the importance of these loss functions in improving the model's performance.

\input{tables/tab3}

\subsection{Embedding Set Element Analysis}
We analyze the role of individual embedding set elements in both \( S^{\mathcal{V}} \) and \( S^{\mathcal{T}} \) modalities in Table \ref{tab:5}. Applying a max operation to pair embeddings from both modalities yields identical results, indicating a robust alignment mechanism. Retaining all embeddings \textit{consistently achieves the best performance}, with the highest RSUM score of $446.53$, highlighting their collective contribution. When using only \( S^{\mathcal{V}} (3) \) or \( S^{\mathcal{T}} (1) \), the RSUM drops to $437.91$, emphasizing the critical role of all components. These findings underscore the advantage of leveraging complete embedding sets, unlike prior methods (e.g., PVSE and SetDiv). The bottom section reports circular variance \(= 1 - \left\| \frac{\sum_{x \in S} x}{|S|} \right\|^2\), where lower $\log$(Var.) values indicate stronger set collapsing, where \ours maintains a higher $\log$(Var.), indicating less set collapse compared to other method, ensuring better disentanglement and a more diverse embedding space.

\input{figures/fig_5}

\subsection{Qualitative Analysis} \label{qualitative_analysis}
To compare the semantic diversity of our learned embeddings with prior work, we perform cross-modal retrieval using each embedding in our embedding set independently. Given a set of embeddings for a query, we retrieve the closest test sample for each query embedding by selecting the best-matched embedding from the test sample’s set. This evaluates how well our embeddings capture diverse semantics. Qualitative results in Figure \ref{fig:qualitative} show that SetDiv representations collapse, retrieving identical or highly similar samples, whereas our embeddings retrieve diverse yet semantically consistent results. The left side of the figure indicates which query embedding was used for retrieval.

%% file: tables/tab2.tex
\begin{table}[ht]
\setlength{\tabcolsep}{2pt}
\centering

\resizebox{\columnwidth}{!}{%
\begin{tabular}{l|c|ccc|ccc|c}
\toprule
\toprule
\multicolumn{2}{c|}{} & \multicolumn{6}{c|}{\textbf{Flickr30k Test Images}} & \\
\cmidrule(lr){0-1}\cmidrule(lr){3-9}
 \multirow{2}{*}{Method}& \multirow{2}{*}{CA}& \multicolumn{3}{c|}{Image-to-Text} & \multicolumn{3}{c|}{Text-to-Image} & \\
 & & R@1 & R@5 & R@10 & R@1 & R@5 & R@10 & RSUM \\
\midrule
\multicolumn{9}{l}{\textit{Resnet152 + Bi-GRU}} \\
\midrule
VSE++    \cite{vse}              & \textcolor{green}{\xmark}  & 52.9      & 80.5      & 87.2      & 39.6      & 70.1      & 79.5      & 409.8  \\
PVSE     \cite{song2019polysemousvisualsemanticembeddingcrossmodal}               & \textcolor{green}{\xmark}  & 59.1      & 84.5      & 91.0        & 43.4      & 73.1      & 81.5      & 432.6  \\
PCME     \cite{chun2021probabilisticembeddingscrossmodalretrieval}               & \textcolor{green}{\xmark}  & 58.5      & 81.4      & 89.3      & 44.3      & 72.7      & 81.9      & 428.1  \\
Set Div   \cite{kim2023setdiverseembeddings}              & \textcolor{green}{\xmark}  & 61.8      & 85.5      & 91.1      & 46.1      & 74.8      & 83.3      & \underline{442.6}  \\
\rowcolor{gray!20}
\ours      & \textcolor{green}{\xmark}  & 68.6      & 89.6      & 94.6      & 51.5     & 78.9     & 86.8     & \textbf{469.9}  \\
\midrule
\multicolumn{9}{l}{\textit{Faster R-CNN + Bi-GRU}} \\
\midrule
SCAN\textsuperscript{\textdagger} \cite{lee2018stackedcrossattentionimagetext}  & \textcolor{red}{\cmark}  & 67.4      & 90.3      & 95.8      & 48.6      & 77.7      & 85.2      & 465    \\
VSRN\textsuperscript{\textdagger} \cite{li2019visual} & \textcolor{green}{\xmark}  & 71.3      & 90.6      & 96.0        & 54.7      & 81.8      & 88.2      & 482.6  \\
CAAN \cite{contextawareattnnw}  & \textcolor{red}{\cmark}  & 70.1      & 91.6      & 97.2      & 52.8      & 79        & 87.9      & 478.6  \\
SGRAF\textsuperscript{\textdagger}    \cite{Diao_Zhang_Ma_Lu_2021} & \textcolor{red}{\cmark}  & 77.8      & 94.1      & 97.4      & 58.5      & 83.0        & 88.8      & 499.6  \\
VSE$\infty$ \cite{chen2021learningbestpoolingstrategy}  & \textcolor{green}{\xmark}  & 76.5      & 94.2      & 97.7      & 56.4      & 83.4      & 89.9      & 498.1  \\
NAAF\textsuperscript{\textdagger}   \cite{negativeawareattnframework_zhang2019}  & \textcolor{red}{\cmark}  & 81.9      & 96.1      & 98.3      & 61.0        & 85.3      & 90.6      & 513.2  \\
CHAN \cite{CHAN2023}  & \textcolor{red}{\cmark}  & 79.7      & 94.5      & 97.3      & 60.2      & 85.3      & 90.7      & 507.8  \\
HREM\textsuperscript{\textdagger} \cite{HREM_2023_CVPR}  & \textcolor{green}{\xmark}  & 81.4      & 96.5      & 98.5      & 60.9      & 85.6      &  91.3      & 514.2  \\
CORA\textsuperscript{\textdagger \textdaggerdbl} \cite{cora2024} & \textcolor{green}{\xmark}   & 82.3      & 96.1     & 98.0      & 63.0      & 87.4      & 92.8        & \textbf{519.6}  \\
Set Div \cite{kim2023setdiverseembeddings}  & \textcolor{green}{\xmark}  & 77.8      & 94.0        & 97.5      & 57.5      & 84.0        & 90.0        & 500.8  \\
\rowcolor{gray!20}
\ours          & \textcolor{green}{\xmark}  & 80.8      & 95.9      & 97.4      & 59.3     & 84.7     & 90.9      & 509.1 \\
\rowcolor{gray!20}
\ours \textsuperscript{\textdagger}        & \textcolor{green}{\xmark}   & 82.1      & 95.6      & 98.3      & 61.6     & 86.3     & 91.9      & \underline{515.8} \\
\midrule
\multicolumn{9}{l}{\textit{Faster R-CNN + BERT}} \\
\midrule
DSRAN\textsuperscript{\textdagger} \cite{Wen_2021_dsran}               & \textcolor{green}{\xmark}  & 77.8      & 95.1      & 97.6      & 59.2      & 86.0        & 91.9      & 507.6  \\
VSE$\infty$ \cite{chen2021learningbestpoolingstrategy}  & \textcolor{green}{\xmark}  & 81.7      & 95.4      & 97.6      & 61.4      & 85.9      & 91.5      & 513.5  \\
MV-VSE\textsuperscript{\textdagger} \cite{Diao_Zhang_Ma_Lu_2021} & \textcolor{green}{\xmark}  & 82.1      & 95.8      & 97.9      & 63.1      & 86.7      & 92.3      & 517.5  \\
CHAN \cite{CHAN2023} & \textcolor{red}{\cmark}  & 80.6      & 96.1      & 97.8      & 63.9      & 87.5      & 92.6      & 518.5  \\
CODER      \cite{wang2022coder}              & \textcolor{red}{\cmark}  & 83.2      & 96.5      & 98.0        & 63.1      & 87.1      & 93.0        & 520.9  \\
HREM\textsuperscript{\textdagger} \cite{HREM_2023_CVPR} & \textcolor{green}{\xmark}  & 84.0      & 96.1        & 98.6      & 64.4      & 88.0      & 93.1      & \underline{524.2}  \\
CORA\textsuperscript{\textdagger \textdaggerdbl} \cite{cora2024} & \textcolor{green}{\xmark}  & 83.4      & 95.9      & 98.6      & 64.1      & 88.1     & 93.1      & 523.3  \\
Set Div \cite{kim2023setdiverseembeddings} & \textcolor{green}{\xmark}  & 81.3      & 95.5        & 97.7      & 62.4     & 86.5     & 91.4     & 514.8 \\
\rowcolor{gray!20}
\ours          & \textcolor{green}{\xmark}  & 84.2      & 96.1      & 97.9      & 63.2     & 87.3     & 92.2    & 520.8 \\
\rowcolor{gray!20}
\ours \textsuperscript{\textdagger} & \textcolor{green}{\xmark}  & 86.2      & 95.7      & 98.4      & 64.8     & 88.8     & 93.2     & \textbf{527.1}
\\
\bottomrule
\bottomrule
\end{tabular}
}
\caption{Recall@K (\%) and RSUM on the Flickr30k dataset. The best RSUM scores are marked in \textbf{bold}, and the second-best scores are \underline{underlined}. CA, ‡, and † indicate models using cross-attention, models that use external data, and ensemble models of two hypotheses.}

\label{table:flickr30k}
\vspace{-1em}

\end{table}

%% file: tables/tab1.tex
\begin{table*}[ht]

\setlength{\tabcolsep}{3pt}
\centering
\begin{adjustbox}{width=\textwidth}
\begin{tabular}{l|c|ccc|ccc|c|ccc|ccc|c}
\toprule
\toprule
\multicolumn{2}{c|} {} & \multicolumn{7}{c|}{\textbf{COCO 1K Test Images}} & \multicolumn{7}{c}{\textbf{COCO 5K Test Images}} \\
\cmidrule(lr){0-1} \cmidrule(lr){3-9} \cmidrule(lr){10-16}
 \multirow{2}{*}{Method}& \multirow{2}{*}{CA} & \multicolumn{3}{c|}{Image-to-Text} & \multicolumn{3}{c|}{Text-to-Image} &  & \multicolumn{3}{c|}{Image-to-Text} & \multicolumn{3}{c|}{Text-to-Image}& \\
 & & R@1 & R@5 & R@10 & R@1 & R@5 & R@10 & RSUM & R@1 & R@5 & R@10 & R@1 & R@5 & R@10 & RSUM  \\
\midrule
\multicolumn{16}{l}{\textit{Resnet152 + Bi-GRU}} \\
\midrule
VSE++ \cite{vse}                & \textcolor{green}{\xmark}  & 64.6      & 90.0        & 95.7      & 52.0        & 84.0        & 92.0        & 478.6  & 41.3      & 71.1      & 81.2      & 30.3       & 59.4      & 72.4     & 355.7   \\
PVSE  \cite{song2019polysemousvisualsemanticembeddingcrossmodal}                & \textcolor{green}{\xmark}  & 69.2      & 91.6      & 96.6      & 55.2      & 86.5      & 93.7      & 492.8  & 45.2      & 74.3      & 84.5      & 32.4       & 63.0        & 75.0       & 374.4   \\
PCME  \cite{chun2021probabilisticembeddingscrossmodalretrieval}                & \textcolor{green}{\xmark}  & 68.8      & -         & -         & 54.6      & -         & -         & -      & 44.2      & -         & -          & 31.9       & -          &     -     &      -   \\
Set Div \cite{kim2023setdiverseembeddings}              & \textcolor{green}{\xmark}  & 70.3      & 91.5      & 96.3      & 56.0        & 85.8      & 93.3      & \underline{493.2}  & 47.2      & 74.8      & 84.1      & 33.8       & 63.1      & 74.7     & \underline{377.7}   \\
\rowcolor{gray!20} 
\ours  & \textcolor{green}{\xmark}  & 74.54     & 93.98     & 97.54     & 58.35     & 87.63     & 94.31     & \textbf{506.75} & 51.84     & 79.86     & 87.62     & 36.35      & 66        & 77.28    & \textbf{398.95}  \\
\bottomrule

\multicolumn{16}{l}{\textit{Faster R-CNN + Bi-GRU}} \\
\midrule
SCAN \textsuperscript{\textdagger} \cite{lee2018stackedcrossattentionimagetext} & \textcolor{red}{\textbf{\cmark}} & 72.7 & 94.8 & 98.4 & 58.8 & 88.4 & 94.8 & 507.9 & 50.4 & 82.2 & 90 & 38.6 & 69.3 & 80.4 & 410.9 \\
VSRN \textsuperscript{\textdagger} \cite{li2019visual} & \textcolor{green}{\textbf{\xmark}} & 76.2 & 94.8 & 98.2 & 62.8 & 89.7 & 95.1 & 516.8 & 53 & 81.1 & 89.4 & 40.5 & 70.6 & 81.1 & 415.7  \\
CAAN \cite{contextawareattnnw} & \textcolor{red}{\textbf{\cmark}} & 75.5 & 95.4 & 98.5 & 61.3 & 89.7 & 95.2 & 515.6 & 52.5 & 83.3 & 90.9 & 41.2 & 70.3 & 82.9 & 421.1 \\
SGRAF\textsuperscript{\textdagger} \cite{Diao_Zhang_Ma_Lu_2021} & \textcolor{red}{\textbf{\cmark}} & 79.6 & 96.2 & 98.5 & 63.2 & 90.7 &  96.1 & 524.3 &  57.8 & 84.9 & 91.6 & 41.9 & 70.7 & 81.3 & 428.2 \\
VSE$_\infty$ \cite{chen2021learningbestpoolingstrategy} & \textcolor{green}{\textbf{\xmark}} & 78.5 & 96 &  98.7 & 61.7 & 90.3 & 95.6 & 520.8 & 56.6 & 83.6 & 91.4 & 39.3 & 69.9 & 81.1 & 421.9 \\
NAAF\textsuperscript{\textdagger} \cite{negativeawareattnframework_zhang2019} & \textcolor{red}{\textbf{\cmark}} & 80.5 &  96.5 & 98.8 &  64.1 &  90.7 & 96.5 & 527.2 &  58.9 & 85.2 & 92.0 & 42.5 & 70.9 & 81.4 & 430.9 \\
CHAN \cite{CHAN2023} & \textcolor{red}{\textbf{\cmark}} & 79.7 & 96.7 &  98.7 & 63.8 & 90.4 & 95.8 & 525.0 & 60.2 & 85.9 &  92.4 & 41.7 & 71.5 & 81.7 &  433.4 \\
HREM \textsuperscript{\textdagger} \cite{HREM_2023_CVPR} & \textcolor{green}{\textbf{\xmark}} & 80.0 & 96.0 &  98.7 & 62.7 & 90.1 & 95.4 & 522.8 & 58.9 & 85.3 & 92.1 & 40.0 & 70.6 & 81.2 & 428.1 \\
CORA \textsuperscript{\textdagger \textdaggerdbl} \cite{cora2024} & \textcolor{green}{\textbf{\xmark}} & 81.7 & 96.7 & 99.0 & 66.0 & 92.0 & 96.7 & \textbf{532.1} & 63.0 & 86.8 & 92.7 & 44.2 &  73.9 & 84.0 & \textbf{444.6} \\
Set Div \cite{kim2023setdiverseembeddings} & \textcolor{green}{\textbf{\xmark}} & 79.8 & 96.2 & 98.6 & 63.6 &  90.7 & 95.7 & 524.6 & 58.8 & 84.9 & 91.5 & 41.1 &  72.0 &  82.4 & 430.7 \\
\rowcolor{gray!20} 
\ours & \textcolor{green}{\textbf{\xmark}} &  80.4 & 96.4 & 98.4 & 64.6 & 91.0 & 96.0 &  526.6 & 59.5 &  85.8 & 94.0 & 42.3 & 72.8 & 83.0 & 436.1 \\
\rowcolor{gray!20} 
\ours \textsuperscript{\textdagger} & \textcolor{green}{\textbf{\xmark}} & 81.7 & 96.5 & 98.7 & 65.4 & 91.6 & 96.5 & \underline{530.3} & 61.9 & 86.86 & 93.1 & 43.1 & 73.8  & 83.9 & \underline{443.1} \\
\bottomrule

\multicolumn{16}{l}{\textit{Faster R-CNN + BERT}} \\
\midrule
VSE$\infty$ \cite{chen2021learningbestpoolingstrategy} & \textcolor{green}{\xmark}  & 79.7      & 96.4      & 98.9      & 64.8      & 91.4      & 96.3      & 527.5  & 58.3      & 85.3      & 92.3      & 42.4       & 72.7      & 83.2     & 434.2   \\
CODER   \cite{wang2022coder}              & \textcolor{red}{\cmark}  & 82.1      & 96.6      & 98.8      & 65.5      & 91.5      & 96.2      & 530.7  & 62.6      & 86.6      & 93.1      & 42.5       & 73.1      & 83.3     & 441.2   \\
MV-VSE\textsuperscript{\textdagger} \cite{Diao_Zhang_Ma_Lu_2021}           & \textcolor{green}{\xmark}  & 80.4      & 96.6      & 99.0        & 64.9      & 91.2      & 96.0        & 528.1  & 59.1      & 86.3      & 92.5      & 42.5       & 72.8      & 83.1     & 436.3   \\
CHAN \cite{CHAN2023} &\textcolor{red}{\textbf{\cmark}}& 81.4      & 96.9      & 98.9      & 66.5      & 92.1      & 96.7      & 532.5  & 59.8      & 87.2      & 93.3      & 44.9       & 74.5      & 84.2     & 443.9   \\
HREM \textsuperscript{\textdagger} \cite{HREM_2023_CVPR} &  \textcolor{green}{\textbf{\xmark}}  & 82.9      & 96.9      & 99.0        & 67.1      & 92.0        & 96.6      & 534.5  & 64.0        & 88.5      & 93.7      & 45.4       & 75.1      & 84.3     & \underline{451.0}     \\
CORA\textsuperscript{\textdagger \textdaggerdbl}  \cite{cora2024}  & \textcolor{green}{\xmark}  & 82.8      & 97.3      & 99.0        & 67.3      & 92.4      & 96.9      & \underline{535.6}  & 64.3      & 87.5      & 93.6      & 45.4       & 74.7      & 84.6     & 450.1   \\
Set Div \cite{kim2023setdiverseembeddings}  & \textcolor{green}{\xmark}  & 82.1     & 95.9      & 95.9      & 65.1     & 91.2      & 96.2     & 529.1  & 62.3      & 86.2     & 92.6     & 42.8      & 72.8     & 82.8    & 439.6  \\
\rowcolor{gray!20} 
\ours       & \textcolor{green}{\xmark}  & 83.0    & 96.9     & 98.9     & 66.4      & 91.9     & 96.6     & 533.8 & 63.3     & 87.9     & 93.2     & 44.2      & 73.9     & 83.9    & 446.5  \\
\rowcolor{gray!20} 
\ours\textsuperscript{\textdagger}       & \textcolor{green}{\xmark}  & 83.8     & 97.0     & 98.9      & 67.3     & 92.4     & 96.9     & \textbf{536.3}   & 65.4     & 88.7     & 93.8     & 45.1     & 74.9     & 84.8    & \textbf{452.6} \\
\bottomrule
\bottomrule
\end{tabular}
\end{adjustbox}
\caption{COCO Recall@K (\%) and RSUM results on both the 1K test setting (average of 5-fold test dataset) and 5K test setting are presented. The best RSUM scores are marked in \textbf{bold}, and the second-best scores are \underline{underlined}. CA, ‡, and † indicate models using cross-attention, models that use external data, and ensemble models, respectively.}

\label{table:mscoco}
\vspace{-1em}
\end{table*}

%% file: tables/tab5.tex
\begin{table}[t]
\resizebox{\columnwidth}{!}{%
\begin{tabular}{c|c|c|c|c|c}
\hline
\hline
\textbf{Similarity} & \textbf{Div} & \textbf{MMD} & \textbf{GD}& \textbf{ISD} & \textbf{RSUM} \\ 
\hline
MIL                     & \cmark & \cmark &        &         &  438.98 \\
Smooth-Chamfer          & \cmark & \cmark &        &         &  439.57 \\
Smooth-Chamfer          & \cmark & \cmark & \cmark &         &  441.76 \\
Smooth-Chamfer          & \cmark & \cmark &        & \cmark  &  442.43 \\
Smooth-Chamfer          &        &        & \cmark & \cmark  &  443.11 \\
Smooth-Chamfer          & \cmark & \cmark & \cmark & \cmark  &  444.10 \\
Maximal Pair Assignment & \cmark & \cmark &        &         &  441.23 \\
Maximal Pair Assignment &        &        & \cmark &         &  442.67 \\
Maximal Pair Assignment &        &        &        & \cmark  &  443.49 \\
Maximal Pair Assignment &        &        & \cmark & \cmark  &  444.49 \\
\rowcolor{gray!20} %
\textbf{Maximal Pair Assignment} & \cmark & \cmark & \cmark & \cmark & \textbf{ 446.53 } \\ 
\hline
\hline
\end{tabular}%
}
\caption{Ablation study of the proposed similarity function and different similarity measures (MIL and Smooth Chamfer) combined with various loss settings, including Div, MMD, GD, and ISD, on the overall performance, measured by RSUM on COCO.}
\label{table_abl}

\end{table}

%% file: tables/tab3.tex
\begin{table}[t]
\resizebox{\columnwidth}{!}{%
\begin{tabular}{cccc|ccc}
\hline
\hline
\multicolumn{4}{c|}{Evaluation} & \multicolumn{3}{c}{RSUM} \\ \hline
$S^\mathcal{V}(1)$ & $S^\mathcal{V}(2)$ & $S^\mathcal{V}(3)$ & $S^\mathcal{V}(4)$ & MIL & Smooth Chamfer & \ours \\ \hline
\rowcolor{gray!20}
\cmark & \cmark & \cmark & \cmark & \textbf{438.98} & \textbf{439.57} & \textbf{4}\textbf{46.53} \\
\cmark &  &  &  & 439.08 & 212.12 & 440.89 \\
 & \cmark &  &  & 389.97 & 439.03 & 443.78 \\
 &  & \cmark &  & 400.06 & 439.50 & 437.91 \\
 &  &  & \cmark & 302.14 & 438.84 & 442.39 \\ \hline
 \hline
$S^\mathcal{T}(1)$ & $S^\mathcal{T}(2)$ & $S^\mathcal{T}(3)$ & $S^\mathcal{T}(4)$ &  &  &  \\ \hline
\rowcolor{gray!20}
\cmark & \cmark & \cmark & \cmark & \textbf{438.98} & \textbf{439.57} & \textbf{446.53} \\
\cmark &  &  &  & 438.84 & 439.24 & 437.91 \\
 & \cmark &  &  & 393.10 & 438.12 & 440.89 \\
 &  & \cmark &  & 293.50 & 438.12 & 442.39 \\
 &  &  & \cmark & 412.32 & 439.14 & 443.78 \\ \hline
 \hline \hline
\multicolumn{4}{c|}{Circular variance of embedding set} & MIL & Smooth Chamfer & \ours \\ \hline
\multicolumn{4}{c|}{ RSUM } & 483.3 & 500.8 & 509.1 \\
\multicolumn{4}{c|}{$\log$(Var.)} & -7.35 & -2.13 & -1.68 \\ \hline
\hline
\end{tabular}%
}
\caption{RSUM on COCO dataset for $S^\mathcal{V}$ (top) and $S^\mathcal{T}$ (middle). This highlights the impact of selectively removing components on overall performance and embedding stability across three methods: MIL, Smooth Chamfer, and \ours. The circular variance of the embedding set (bottom) on Flickr30k, where lower $\log$(Var.) values indicate stronger collapse.}

\label{tab:5}

\end{table}

%% file: figures/fig_5.tex
\begin{figure}[h!]
  \centering
  \vspace{-0.8em}
  \includegraphics[width=1\linewidth]{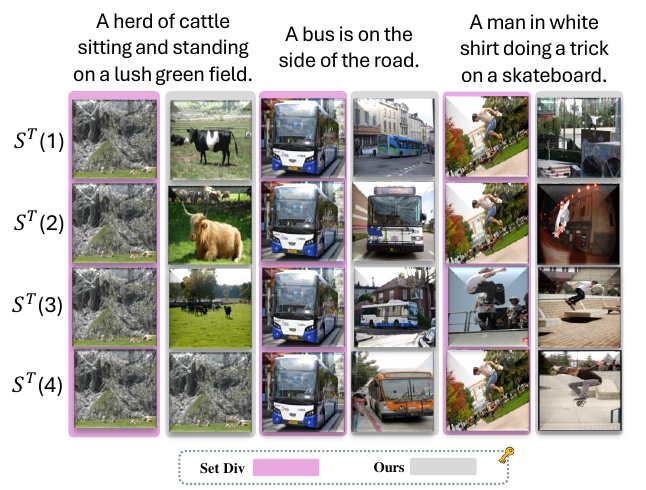}
\vspace{-2em}
   \caption{Visual comparison of image retrieval results between our method and Set Div. For each embedding in the set, we show the top retrieved image. Our method generates embeddings that retrieve diverse images, while Set Div's embeddings tend to collapse, resulting in retrieval of identical or highly similar images.}
   \label{fig:qualitative}
\end{figure}

%% file: sec/5_conclusion.tex
\vspace{-0.5em}

\section{Conclusion}
We introduced a novel set-based embedding framework for cross-modal retrieval with Maximal Pair Assignment Similarity and a combination of new loss functions. The Maximal Pair Assignment Similarity, utilizing permutation-based assignments, resolves challenges of sparse supervision and set collapsing, improving accurate image-text similarity scoring. Global Discriminative Loss further enhances the model's ability to differentiate between embeddings, while the Intra-Set Divergence Loss mitigates set collapsing by encouraging diversity within each set. Our approach achieves superior performance on standard benchmarks compared to previous methods, and we plan to extend it to other modalities and tasks in future work.

\section{Acknowledgements}

We acknowledge Advanced Research Computing at Virginia Tech for providing computational resources and technical support that have contributed to the results reported within this paper. We also thank all reviewers for their comments which helped improve the paper.

\section{Limitation}
While our proposed \ours demonstrates strong performance for cross-modal retrieval and overcomes many important limitations related to representation set collapse, several important limitations remain unaddressed. While MS-COCO and Flickr30k are standard benchmarks, they represent a relatively constrained subset of real-world cross-modal scenarios. Most image captions in these datasets focus on describing concrete objects and actions, rather than abstract concepts, emotional content, or diverse cultural interpretations that might benefit from multiple distinct representations. Future work should explore more diverse and challenging datasets that better reflect the complexity of human visual and linguistic understanding. In addition , while effective for image-text pairs, the current formulation of MaxMatch is specific to bi-modal retrieval. Extending the framework to handle additional modalities or simultaneous alignment across three or more modalities would require substantial modifications to both the matching mechanism and loss functions. This limits its immediate applicability in broader multimodal scenarios involving audio, video, or other data types. These limitations point to several promising directions for future research, including more efficient matching algorithms, adaptive set size mechanisms, and generalization to broader multimodal contexts.

%% file: sec/supp.tex
\newpage

\section{Appendix}
This Appendix provides additional details and results that complement the main paper. Section \ref{Implementation} outlines our comprehensive implementation for various model configurations. In Section \ref{smooth-chamfer-collapse}, we show that smooth Chamfer similarity encourages set collapse. Methodological considerations regarding external knowledge integration are discussed in Section \ref{CORA}. Section \ref{Larger_Backbone} evaluates our approach using stronger visual backbones, and Section \ref{PVSE} examines its performance on the PVSE architecture. Finally, Section \ref{Code} presents pseudocode for our key components.

\vspace{1em}

\section{Implementation Details} \label{Implementation}
Following prior work \cite{kim2023setdiverseembeddings, song2019polysemousvisualsemanticembeddingcrossmodal}, we use two types of feature extractors and set the embedding dimension $D = 1024$. We obtain convolutional visual features by applying a $1 \times 1$ convolution to the final feature map of a CNN. Then, we use the pre-extracted 2048-dimensional region features from Faster R-CNN \cite{faster_rcnn} as BUTD \cite{anderson2018bottomuptopdownattentionimage}. Both feature types were transformed to the embedding space via a two-layer MLP with residual connections. We set $K=4$ and $D_h=1024$ for convolutional features and $D_h=2048$ for region features \cite{kim2023setdiverseembeddings}. We process textual features using a BiGRU with either GloVe \cite{glove} or BERT \cite{devlin2019bertpretrainingdeepbidirectional} embeddings.

Our implementation uses PyTorch \cite{paszke2017automatic} v2.0.1, leveraging the model architecture from the SetDiv codebase \cite{kim2023setdiverseembeddings}.
We employ automatic mixed precision to improve training speed and efficiency.
Each input batch consists of either 128 or 200 images along with their corresponding captions.
We train our model using the AdamW optimizer, with both MMD loss ($\lambda_{MMD}$) and diversity loss ($\lambda_{Div}$) weighted at 0.01 across all configurations.
While the main paper describes the key implementation details, specific training parameters vary based on the feature extractors used.

\vspace{.5em}
\noindent \textbf{ResNet152 + BiGRU:} 
We train the model for 120 epochs using initial learning rates of 1e-3 for MS-COCO \cite{cococommonobjects} and 2e-3 for Flickr30k \cite{flickr30k}.
For the set prediction module, these learning rates are scaled by factors of 0.1 and 0.01 for MS-COCO and Flickr30k, respectively.
Following \cite{song2019polysemousvisualsemanticembeddingcrossmodal, kim2023setdiverseembeddings}, we apply step-wise learning rate decay with a factor of 0.1 every 20 epochs.
During the first 50 epochs, the CNN weights are kept frozen.
Training is performed on dual NVIDIA A40 GPUs with a batch size of 200.
For the loss configuration, the triplet margin ($\delta_1$) is set to 0.2, while both Global Discriminative Loss ($\lambda_{GD}$) and Intra-Set Divergence Loss ($\lambda_{ISD}$) use weights of 0.1.
The margin parameters ($\delta_{2,3}$) and scaling parameter ($s$) are set to 0.6 and 0.5, respectively. 
Contrastive loss is not included in this configuration.

\vspace{.5em}
\noindent \textbf{Faster R-CNN + BiGRU:} 
The model is trained for 80 epochs with an initial learning rate of 1e-3, employing cosine annealing \citep{loshchilov2016sgdr} and weight decay of 1e-4.
For the set prediction module, the learning rate is scaled by 0.1 for both datasets.
Following \cite{chen2021learningbestpoolingstrategy}, we apply a dropout of 20\% to both ROI features and word embeddings during training.
Training is conducted on a single NVIDIA A40 GPU with a batch size of 200.
For the loss configuration, the triplet margin ($\delta_1$) is set to 0.3, while both Global Discriminative Loss ($\lambda_{GD}$) and Intra-Set Divergence Loss ($\lambda_{ISD}$) use weights of 0.1 for MS-COCO and 0.05 for Flickr30K.
The margin parameters ($\delta_{2,3}$) and scaling parameter ($s$) are set to 0.6 and 0.5, respectively.
The contrastive loss weight is set to 0.001.

\vspace{.5em}
\noindent \textbf{Faster R-CNN + BERT:} 
The model trains for 45 epochs with an initial learning rate of 1e-3, employing cosine annealing and weight decay of 5e-4. 
For the set prediction module and BERT, the learning rates are scaled by 0.5 and 0.1, respectively.
Following previous configurations, we apply dropout of 20\% to both ROI features and word embeddings during training.
Training is conducted on dual NVIDIA A40 GPUs with batch sizes of 200 for MS-COCO and 128 for Flickr30k.
For the loss configuration, the triplet margin ($\delta_1$) is set to 0.15, while both Global Discriminative Loss ($\lambda_{GD}$) and Intra-Set Divergence Loss ($\lambda_{ISD}$) use weights of 0.1.
The margin parameters ($\delta_{2,3}$) and scaling parameter ($s$) are set to 0.8 and 0.5, respectively.
The contrastive loss weight is set to 0.001.

\newpage

\section{Smooth Chamfer Leads to Set Collapse}
\label{smooth-chamfer-collapse}

We consider the smooth Chamfer similarity defined as
\begin{multline}
S_{SC}(S_1,S_2) = \frac{1}{2\alpha |S_1|} \sum_{x \in S_1} \log \Biggl( \sum_{y \in S_2} e^{\alpha c(x,y)} \Biggr) \\
+ \frac{1}{2\alpha |S_2|} \sum_{y \in S_2} \log \Biggl( \sum_{x \in S_1} e^{\alpha c(x,y)} \Biggr).
\end{multline}
In what follows, we show, under suitable assumptions (e.g. when $c(x,y)$ is affine in $x$, as in $c(x,y)=x^\top y$), that the objective is minimized when all elements in each set are identical. For clarity, we detail the argument for $S_1$, and a symmetric argument applies to $S_2$.

Let 
$S_1 = \{x_1, x_2, \dots, x_n\}$.
Assume that $S_2$ is fixed. Define the function $
L(x) \triangleq \log\Biggl( \sum_{y\in S_2} e^{\alpha\, c(x,y)} \Biggr)$. Then, the contribution of $S_1$ to the overall objective is $ f(S_1) = \frac{1}{2\alpha n} \sum_{i=1}^n L(x_i)$.
(We omit the constant prefactor $1/(2\alpha)$ in what follows, as it does not affect the location of the minimum.)

Assume that $c(x,y)$ is affine in $x$ for every fixed $y$; for example, if $c(x,y)= x^\top y,$
then for fixed $y$ the map $x\mapsto \alpha\, c(x,y)$ is affine. Since the exponential is convex and increasing, the composition $x\mapsto e^{\alpha\, c(x,y)}$ is convex. Moreover, the sum over $y$ of convex functions is convex, and the logarithm of a sum of exponentials (the log-sum-exp function) is also convex. Thus, the function $L(x) = \log\Biggl(\sum_{y\in S_2} e^{\alpha\, c(x,y)}\Biggr)$ is convex in $x$.

Let the arithmetic mean of the embeddings in $S_1$ be $\bar{x} = \frac{1}{n}\sum_{i=1}^n x_i$.
Then by Jensen's inequality we have:
\begin{equation}
\frac{1}{n}\sum_{i=1}^n L(x_i) \geq L\Biggl(\frac{1}{n}\sum_{i=1}^n x_i\Biggr) = L(\bar{x}) \;.
\end{equation}
Equality holds if and only if all $x_i$ are equal. That is, for any fixed mean $\bar{x}$, the minimum of
\begin{equation}
f(S_1)=\frac{1}{n}\sum_{i=1}^n L(x_i)
\end{equation}
is achieved if and only if
\begin{equation}
x_1=x_2=\cdots=x_n=\bar{x}.
\end{equation}
Thus, any deviation (i.e., any diversity among the $x_i$) increases the objective.

This also becomes obvious when analyzed from the perspective of the gradient. Consider the gradient of $L(x)$ with respect to $x$. By the chain rule,
\begin{equation} 
\nabla_x L(x) = \frac{\sum_{y\in S_2} \alpha\, e^{\alpha\, c(x,y)}\, \nabla_x c(x,y)}{\sum_{y\in S_2} e^{\alpha\, c(x,y)}}.
\end{equation}

Thus, each $x_i$ receives a gradient that is a softmax-weighted average of the vectors $\alpha\, \nabla_x c(x,y)$ (for example, when $c(x,y)=x^\top y$, one has $\nabla_x c(x,y)=y$). 

Now, consider a small perturbation of the set $S_1$ that preserves the overall mean.
\begin{equation}
x_i = \bar{x} + \delta_i,\qquad \text{with}\qquad \sum_{i=1}^n \delta_i = 0 .
\end{equation}
Because $L$ is convex, its Hessian denoted by
\begin{equation}
H(x) = \nabla^2_x L(x),
\end{equation}
is positive semidefinite:
\begin{equation}   
H(x) \succeq 0.
\end{equation}
A standard result in convex analysis is that the average
\begin{equation}
\frac{1}{n}\sum_{i=1}^n L(\bar{x}+\delta_i)
\end{equation}
is minimized (with respect to the deviations $\{\delta_i\}$ under the constraint $\sum_i \delta_i=0$) when all $\delta_i=0$. In other words, any nonzero differences $\delta_i$ (i.e., any diversity among the embeddings) increases the objective's value. Hence, the Hessian is strictly positive in the directions that would cause dispersion in $S_1$. Thus, the gradients computed above will push any $x_i$ that deviates from the mean $\bar{x}$ back toward $\bar{x}$. Hence, during training the optimal (or at least a stationary) point is one where all embeddings in $S_1$ are equal. A symmetric argument applies to $S_2$.

\section{Discussion on CORA and 3SHNet External Knowledge Integration} \label{CORA}

Recent work by CORA \cite{cora2024} achieves impressive results through the integration of an external LLM as a scene graph parser \cite{li2023factual}, particularly when using Bi-GRU encoders. While this approach demonstrates strong performance, it introduces important considerations for fair comparison. CORA's key innovation lies in enhancing text representations by augmenting raw captions with structured scene graphs derived from an LLM parser, effectively incorporating external semantic knowledge into the representation learning process. This external knowledge particularly benefits simpler text encoders like Bi-GRU, where the additional semantic structure helps bridge the gap in language understanding capabilities. CORA achieves state-of-the-art results even with embeddings trained from scratch (as reported in their supplementary material), highlighting how the LLM parser significantly improves textual representations.

In contrast, 3SHNet \cite{ge20243shnet} takes a different approach by integrating external visual knowledge through semantic segmentation features. Rather than relying on a language-based external module, 3SHNet incorporates structured spatial and object-level information using UPSNet to enhance object-region representations. This structured semantic-spatial self-highlighting method improves image representation, making retrieval more robust without requiring external textual scene graphs.

These approaches raise important methodological considerations. Unlike CORA and 3SHNet, our method operates directly on raw captions using a standard text encoder (GloVe or BERT) without incorporating external knowledge or additional preprocessing, and does not rely on UPSNet segmentation beyond the basic steps used in prior work \cite{song2019polysemousvisualsemanticembeddingcrossmodal, kim2023setdiverseembeddings, HREM_2023_CVPR, vse, chen2021learningbestpoolingstrategy}. This distinction becomes particularly evident when using stronger language models like BERT, where the benefits of external parsing diminish as the encoder itself becomes more capable of understanding complex semantic relationships. While CORA's paper acknowledges that BERT excels at encoding longer text sequences, their approach primarily uses it to encode short phrases, which loses the global context of the sentence. Our method, in contrast, leverages BERT's full capability to process complete captions, achieving competitive performance without requiring additional scene graph parsing or external knowledge integration. This demonstrates the effectiveness of our approach in maintaining semantic richness while keeping the architecture simpler and more efficient.

\section{Testing with a Larger Backbone} \label{Larger_Backbone}
In alignment with prior work \cite{kim2023setdiverseembeddings}, we select ResNeXT101 \cite{xie2017aggregated} pretrained on the Instagram dataset \cite{li2019visual} as the larger visual extractor paired with BERT \cite{devlin2019bertpretrainingdeepbidirectional}.
The model trains for 50 epochs with an initial learning rate of 1e-4, which decays by a factor of 0.1 every 20 epochs.
Consistent with \cite{vse}, the CNN's learning rate is scaled down by 0.1, and the CNN remains frozen during the first epoch. 
During this initial phase, we employ triplet loss without hard negative mining, then transition to hardest negative mining in subsequent epochs.
For the loss configuration, both Global Discriminative Loss ($\lambda_{GD}$) and Intra-Set Divergence Loss ($\lambda_{ISD}$) use weights of 0.1.
The margin parameters ($\delta_{2,3}$) and scaling parameter ($s$) are set to 0.6 and 0.5, respectively, with a triplet margin ($\delta_1$) of 0.1. 
Contrastive loss is not included in this configuration.
Training is performed on dual A100 PCIe GPUs with a batch size of 128.
This visual backbone is also used in VSE \cite{chen2021learningbestpoolingstrategy} and Set Div \cite{kim2023setdiverseembeddings}. The results on the MS-COCO and Flickr30K test sets are displayed in Table \ref{tab:resnext}, demonstrating performance improvements with our method. 

The results in Table \ref{tab:resnext} demonstrate the effectiveness of our method when combined with stronger visual backbones.
On Flickr30K, our method achieves significant improvements over both VSE$\infty$ and Set Div baselines, with notable gains in Image-to-Text retrieval (R@1 improves from 88.8\% to 90.5\%).
Similar improvements are observed on MS-COCO, where our method consistently outperforms previous approaches across both 1K and 5K test sets.
Particularly in the more challenging 5K setting, we improve the RSUM from 474.9 to 475.14, demonstrating our method's robustness to larger retrieval spaces.

\input{tables/tab_resnext}

\input{tables/tab_PVSE}

\section{Evaluating Method Performance on PVSE}\label{PVSE}

We evaluate the versatility and effectiveness of our proposed method by applying it to the PVSE architecture, demonstrating its ability to enhance various baseline models.
While PVSE offers more modest performance compared to recent methods like SDE, it provides an excellent test case for assessing our method's generalizability.
The results in Table \ref{tab:PVSE} show substantial improvements across all metrics when our method is integrated with PVSE.
Specifically, our method improves PVSE's Image-to-Text R@1 performance from 59.1\% to 65.4\%, a significant gain of 6.3 percentage points, surpassing even the more recent Set Div model (61.8\%).
In Text-to-Image retrieval, we achieve consistent improvements, with R@1 increasing from 43.4\% to 45.4\%.
The overall RSUM metric improves by 18 points (from 432.6 to 450.6), demonstrating the comprehensive enhancement our method brings to the baseline model.

Notably, our method's performance with PVSE not only surpasses the original PVSE results but also outperforms several more recent approaches including PCME and approaches the performance of Set Div.
This is particularly impressive given PVSE's relatively simpler architecture.
A key distinction of our approach is its ability to utilize the entire embedding set during training and inference, whereas traditional Multiple Instance Learning (MIL) approaches in PVSE only select a single embedding from the set. This comprehensive utilization enables our method to better capture diverse nuanced meanings from the samples, leading to more robust representations.
When further combined with Set Div, our method achieves even more substantial gains, reaching state-of-the-art performance with an RSUM of 469.9, highlighting its complementary nature to existing advanced techniques. 
These results convincingly demonstrate that our method is architecture-agnostic and can effectively enhance both simple and sophisticated models, validating its broad applicability in cross-modal retrieval tasks.

\section{Qualitative Analysis for Image-to-Text Retrieval}

\begin{figure}[t]
  \centering
  
  \includegraphics[width=1\linewidth]{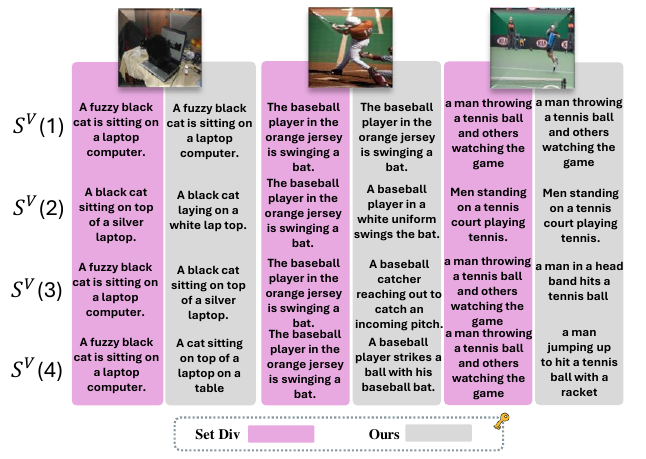}

   \caption{Visual comparison of image retrieval results between our method and Set Div. For each embedding in the set, we show the top retrieved image. Our method generates embeddings that retrieve diverse images, while Set Div's embeddings tend to collapse, resulting in retrieval of identical or highly similar images.}
   \label{fig:qualitativei2t}
\end{figure}

Following Section \ref{qualitative_analysis}, we extend our qualitative analysis to image-to-text retrieval. We perform cross-modal retrieval using each embedding in our set independently, selecting the closest test sample for each query embedding. This evaluates the semantic diversity captured by our embeddings compared to prior work. As shown in Figure \ref{fig:qualitativei2t}, SetDiv representations tend to collapse, retrieving nearly identical samples, whereas our embeddings yield diverse yet semantically coherent results. The left side of the figure indicates the query embedding used for retrieval.  

\begin{figure}[t]
  \centering
  \includegraphics[width=1\linewidth]{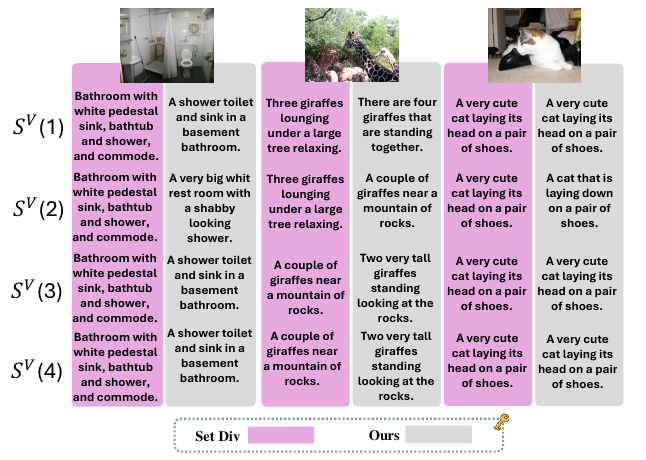}
  \caption{Failure cases in image-to-text retrieval using our method. While our approach typically retrieves semantically diverse captions, some examples particularly those with visually homogeneous content result in repeated or overly similar captions across the set. This highlights a limitation in capturing fine-grained diversity for certain input images.}
  \label{fig:negative_samples}
\end{figure}

\noindent \textbf{Failure Cases and Limitations:}
While our method generally produces more semantically diverse and accurate retrievals compared to SetDiv, we also observe certain failure cases where diversity is reduced or retrievals collapse. As shown in the rightmost column of Figure~\ref{fig:negative_samples}, our embeddings occasionally yield highly similar or even repeated captions across different slots, especially for images with limited semantic variation or dominant foreground objects. This suggests that, in scenarios where visual content is overly focused or lacks contextual richness, the learned embeddings may converge to similar representations. To further characterize these limitations, we include additional qualitative examples in the appendix, including cases where our model retrieves semantically plausible yet not part of the ground truth captions, a common occurrence in image-text datasets.

\section{Code Availability} \label{Code}
We provide pseudocodes of our key components illustrating the Global Discriminative Loss, Intra-Set Divergence Loss, and Maximal Pair Assignment Similarity functions. These code snippets demonstrate the conceptual implementation of our method in PyTorch notation, highlighting both the efficiency and simplicity of our approach. We will release the codebase on GitHub. 

\newpage

\section*{Maximal Pair Assignment Similarity}
\vspace{-.5em}
\begin{minipage}{0.98\linewidth}
\begin{lstlisting}[language=Python, caption={Generate a mask for maximum similarity pairs in a similarity matrix.}, label={lst:mask_similarity}]
def mask_max_similarity(sim_matrix, text_indices, img_indices):
    """
    Generate a binary mask to identify maximum similarity pairs in a similarity matrix.

    Args:
        sim_matrix (Tensor): Similarity matrix between text and image embeddings.
        text_indices (Tensor): Indices corresponding to text embeddings.
        img_indices (Tensor): Indices corresponding to image embeddings.

    Returns:
        Tensor: Binary mask with 1s at maximum similarity pairs.
    """
    # Step 1: Extract similarities at specified indices
    selected_similarities = sim_matrix[text_indices, img_indices]

    # Step 2: Find indices of maximum similarity for each pair
    max_indices = selected_similarities.argmax(dim=1)

    # Step 3: Create binary mask with 1s at maximum similarity indices
    mask = torch.zeros_like(sim_matrix)
    mask[text_indices[max_indices], img_indices[max_indices]] = 1

    return mask
\end{lstlisting}

\vspace{-1em}

\begin{lstlisting}[language=Python, caption={Create index permutations for matrix operations.}, label={lst:create_permutations}]
def create_index_permutations(num_embeddings, row_size, col_size):
    # Generate permutations of indices for matrix operations.
    # Step 1: Generate all permutations of the embeddings
    permutations = torch.tensor(list(itertools.permutations(range(num_embeddings))), dtype=torch.long)
    # Step 2: Repeat permutations to match the required row size
    row_indices = permutations.repeat(row_size // num_embeddings, 1)
    # Step 3: Create column indices to match the required column size
    col_indices = torch.arange(col_size).repeat(row_indices.size(0), 1)
    # tuple: Two tensors representing row and column indices.
    return row_indices, col_indices
\end{lstlisting}
\end{minipage}

\begin{lstlisting}[language=Python, caption={Function to compute assignment of maximum similarity between embeddings.}, label={lst:assignment_similarity}]
def maximal_pair_assignment_similarity(img_embs, txt_embs):
    """
    Compute assignment of maximum similarity between image and text embeddings.

    Args:
        img_embs (Tensor): Image embeddings.
        txt_embs (Tensor): Text embeddings.

    Returns:
        Tensor: Maximum similarity tensor.
    """
    # Step 1: Compute cosine similarity between normalized embeddings
    dist = cosine_sim(l2norm(img_embs), l2norm(txt_embs))

    # Step 2: Determine row and column sizes based on batch and set sizes
    row_size = image_batch_size * img_set_size  # Number of image embeddings
    col_size = text_batch_size * txt_set_size  # Number of text embeddings

    # Step 3: Generate all possible index permutations for matching
    text_index_all, image_index_all = create_index_permutations(img_set_size, row_size, col_size)

    # Step 4: Create a mask for maximum similarity pairs
    mask = mask_max_similarity(dist.detach(), text_index_all, image_index_all)

    # Step 5: Calculate maximum similarity scores
    max_similarity = mask * dist

    # Step 6: Apply exponential scaling and average pooling
    max_similarity = avg_pool(torch.exp(max_similarity.unsqueeze(0)) - 1) * img_set_size

    return max_similarity.squeeze()
\end{lstlisting}

\newpage

\section*{Intra-Set Divergence Loss}

\begin{lstlisting}[language=Python, caption={Intra-Set Divergence Loss}]
def Intra_Set_Divergence_Loss(set_emb, margin, scale):        
    # Compute pairwise similarity between all embeddings in the set
    A = torch.bmm(set_emb, set_emb.transpose(1, 2))  # [bs, K, K]
    
    # Create a mask to exclude self-similarities and include only upper triangular matrix elements
    num_embeddings = set_emb.size(1)
    mask = torch.triu(torch.ones(num_embeddings, num_embeddings, device=set_emb.device), diagonal=1).bool()
    mask = mask.unsqueeze(0)  # Expand mask to match batch dimension
    
    # Apply the mask to extract unique pairwise similarities
    A_masked = A.masked_select(mask)
    
    # Compute the loss using the smooth exponential function with margin and scale
    loss = torch.exp((A_masked - margin) * scale)
    
    # Normalize the loss over the batch and number of unique pairs
    return loss.sum() / (set_emb.shape[0] * (set_emb.shape[1] * (set_emb.shape[1] - 1) / 2))
\end{lstlisting}

\section*{Global Discriminative Loss}

\begin{lstlisting}[language=Python, caption={Global Discriminative Loss}, label={lst:GD}]
def Global_Discriminative_Loss(Set_emb, global_emb, margin, scale):        
    # Add a singleton dimension to global_embedding for batch matrix multiplication
    global_emb= global_emb.unsqueeze(1) #[bs,1,D]

    # Transpose embeddings to match dimensions for batch matrix multiplication
    Set_emb= Set_emb.transpose(1, 2) #[bs,D,K]
    
    # Compute pairwise similarity between global embedding and the set of embeddings
    A= torch.bmm(global_emb, Set_emb).squeeze() #[bs,K]
    
    # Compute the loss by applying the margin and scale, then take the mean
    loss= torch.mean(torch.exp(scale * (A - margin)))
    return loss
\end{lstlisting}

\newpage

%% file: tables/tab_resnext.tex
\begin{table}[t]
\resizebox{\columnwidth}{!}{%
\begin{tabular}{lccccccc}
\hline
\multicolumn{1}{l|}{} & \multicolumn{3}{c|}{Image-to-Text} & \multicolumn{3}{c|}{Text-to-Image} & \multicolumn{1}{l}{\multirow{2}{*}{RSUM}} \\ \cline{1-7}
\multicolumn{1}{l|}{} & R@1 & R@5 & \multicolumn{1}{c|}{R@10} & R@1 & R@5 & \multicolumn{1}{c|}{R@10} & \multicolumn{1}{l}{} \\ \hline
\multicolumn{8}{l}{MS-COCO 1K Test Images} \\ \hline
\multicolumn{1}{l|}{VSE$\infty$} & 84.5 & 98.1 & \multicolumn{1}{c|}{99.4} & 72 & 93.9 & \multicolumn{1}{c|}{97.5} & 545.4 \\
\multicolumn{1}{l|}{Set Div} & 86.3 & 97.8 & \multicolumn{1}{c|}{99.4} & 72.4 & 94 & \multicolumn{1}{c|}{97.6} & \underline{547.5} \\
\rowcolor{gray!20} 
\multicolumn{1}{l|}{\ours} & 86.9 & 98.12 & \multicolumn{1}{c|}{99.38} & 72.22 & 94.12 & \multicolumn{1}{c|}{97.6} & \textbf{547.84} \\ \hline
\multicolumn{8}{l}{MS-COCO 5K Test Images} \\ \hline
\multicolumn{1}{l|}{VSE$\infty$} & 66.4 & 89.3 & \multicolumn{1}{c|}{94.6} & 51.6 & 79.3 & \multicolumn{1}{c|}{87.6} & 468.9 \\
\multicolumn{1}{l|}{Set Div} & 69.1 & 90.7 & \multicolumn{1}{c|}{95.6} & 52.1 & 79.6 & \multicolumn{1}{c|}{87.8} & \underline{474.9} \\
\rowcolor{gray!20} 
\multicolumn{1}{l|}{\ours} & 69.62 & 90.76 & \multicolumn{1}{c|}{95.54} & 52.18 & 79.46 & \multicolumn{1}{c|}{87.58} & \textbf{475.14} \\ \hline
\multicolumn{8}{l}{Flickr30k Test Images} \\ \hline
\multicolumn{1}{l|}{VSE$\infty$} & 88.4 & 97.3 & \multicolumn{1}{c|}{99.5} & 74.2 & 93.7 & \multicolumn{1}{c|}{96.8} & 550.9 \\
\multicolumn{1}{l|}{Set Div} & 88.8 & 98.5 & \multicolumn{1}{c|}{99.6} & 74.3 & 94 & \multicolumn{1}{c|}{96.7} & \underline{551.9} \\
\rowcolor{gray!20} 
\multicolumn{1}{l|}{\ours} & 90.5 & 99 & \multicolumn{1}{c|}{99.8} & 74.96 & 94.2 & \multicolumn{1}{c|}{96.78} & \textbf{555.24} \\ \hline
\end{tabular}%
}
\caption{Recall@K (\%) and RSUM on MS-COCO and Flickr30k dataset with ResNeXT101 and BERT backbones. The best RSUM scores are marked in \textbf{bold}, and the second-best scores are \underline{underlined}.}
\label{tab:resnext}
\end{table}

%% file: tables/tab_PVSE.tex
\begin{table}[t]
\resizebox{\columnwidth}{!}{%
\begin{tabular}{cccccccc}
\hline
\multicolumn{1}{l|}{} & \multicolumn{3}{c|}{Image-to-Text} & \multicolumn{3}{c|}{Text-to-Image} & \multicolumn{1}{l}{\multirow{2}{*}{RSUM}} \\ \cline{1-7}
\multicolumn{1}{l|}{} & R@1 & R@5 & \multicolumn{1}{c|}{R@10} & R@1 & R@5 & \multicolumn{1}{c|}{R@10} & \multicolumn{1}{l}{} \\ \hline
\multicolumn{1}{c|}{VSE++} & 52.9 & 80.5 & \multicolumn{1}{c|}{87.2} & 39.6 & 70.1 & \multicolumn{1}{c|}{79.5} & 409.8 \\
\multicolumn{1}{c|}{PVSE} & 59.1 & 84.5 & \multicolumn{1}{c|}{91} & 43.4 & 73.1 & \multicolumn{1}{c|}{81.5} & 432.6 \\
\multicolumn{1}{c|}{PCME} & 58.5 & 81.4 & \multicolumn{1}{c|}{89.3} & 44.3 & 72.7 & \multicolumn{1}{c|}{81.9} & 428.1 \\
\multicolumn{1}{c|}{Set Div} & 61.8 & 85.5 & \multicolumn{1}{c|}{91.1} & 46.1 & 74.8 & \multicolumn{1}{c|}{83.3} & 442.6 \\
\rowcolor{gray!20} 
\multicolumn{1}{l|}{\textbf{\ours + PVSE} } & \multicolumn{1}{c}{\textbf{65.4}} & \multicolumn{1}{c}{\textbf{88.1}} & \multicolumn{1}{c|}{\textbf{93.7}} & \multicolumn{1}{c}{\textbf{45.4}} & \multicolumn{1}{c}{\textbf{74.7}} & \multicolumn{1}{c|}{\textbf{83.2}} & \multicolumn{1}{c}{\textbf{450.6}} \\
\rowcolor{gray!20}
\multicolumn{1}{l|}{\ours} & 68.6 & 89.6 & \multicolumn{1}{c|}{94.6} & 51.5 & 78.9 & \multicolumn{1}{c|}{86.8} & 469.9 \\ \hline
\end{tabular}%
}
\caption{Recall@K (\%) and RSUM on Flickr30k dataset on PVSE architicture. The scores for PVSE with our method is in \textbf{bold}.}
\label{tab:PVSE}
\end{table}